\pdfoutput=1
\documentclass[10pt,twocolumn]{article}

\usepackage[letterpaper,top=0.75in,bottom=0.85in,left=0.75in,right=0.75in,columnsep=0.3in]{geometry}
\usepackage[T1]{fontenc}
\usepackage[utf8]{inputenc}
\usepackage{mathptmx}          
\usepackage[scaled=0.92]{helvet}
\usepackage{courier}           
\usepackage{microtype}
\usepackage{amsmath,amssymb}
\usepackage{graphicx}
\usepackage{booktabs}
\usepackage{tabularx}
\usepackage{multirow}
\usepackage{array}
\newcolumntype{L}[1]{>{\raggedright\arraybackslash}p{#1}}
\newcolumntype{Y}{>{\raggedright\arraybackslash}X}
\usepackage[table]{xcolor}
\usepackage{enumitem}
\usepackage{caption}
\usepackage{dblfloatfix}
\usepackage{tikz}
\usetikzlibrary{positioning,arrows.meta,fit,backgrounds,calc}
\usepackage{pgfplots}
\pgfplotsset{compat=1.16}
\usepackage{algorithm}
\usepackage{algpseudocode}
\usepackage[numbers,sort&compress]{natbib}
\PassOptionsToPackage{hyphens}{url}
\usepackage[colorlinks=true,linkcolor=blue!50!black,citecolor=blue!50!black,urlcolor=blue!50!black]{hyperref}
\usepackage[capitalise,noabbrev]{cleveref}
\hypersetup{pdftitle={FinVeriQA: Financial Research Agents for Question Answering: Routing, Retrieval, MCP Tool Use, and Provenance-Aware Evidence Verification},
  pdfauthor={Meenakshi Rajpurohit, Anna Phan, Ravikumar Komandur Narayanan, Gunanidhi Ramakrishnan, Nathan Howland}}

\definecolor{laycol}{HTML}{E39B2D}   
\definecolor{miscol}{HTML}{2F5D8A}   
\definecolor{pycol}{HTML}{0E7C66}    
\definecolor{errcol}{HTML}{C4483E}   
\definecolor{gptcol}{HTML}{6A4FA0}   
\definecolor{minicol}{HTML}{6C93BF}  
\definecolor{slatecol}{HTML}{8795A6} 

\newcommand{\sys}{FinVeriQA}

\newcommand{\code}[1]{\texttt{#1}}

\setlist{itemsep=1pt,topsep=3pt,leftmargin=*}

\title{\textbf{FinVeriQA: Financial Research Agents for Question Answering:\\
Routing, Retrieval, MCP Tool Use, and\\
Provenance-Aware Evidence Verification}}

\author{%
Meenakshi Rajpurohit\textsuperscript{1} \qquad
Anna Phan\textsuperscript{1} \qquad
Ravikumar Komandur Narayanan\textsuperscript{1}\\
Gunanidhi Ramakrishnan\textsuperscript{1} \qquad
Nathan Howland\textsuperscript{1}\\[6pt]
\textsuperscript{1}San Jos\'e State University, San Jos\'e, CA, USA}

\date{November 2026}

\begin{document}
\maketitle

\begin{abstract}
Large language models (LLMs) write fluent financial analyses but miscompute ratios, recall stale prices, and route questions to the wrong tool with high confidence. We ask which component should do each job in a financial question-answering assistant, and study it with \sys{}, an open-source multi-agent research prototype over SEC filings and live market data: a compact typed-decision model (Laya) routes each question and screens out advice requests, Python computes every number, retrieval and tools fetch evidence, Mistral-7B writes a cited answer, a code validator checks that cited IDs exist and that numbers match computed notes, and a Model Context Protocol (MCP) server exposes the pipeline. In live experiments, Laya routed better than Mistral-7B (0.85 vs.\ 0.52; overall 0.86 vs.\ 0.68, McNemar $p = 0.0003$); a 12B Mistral decided no better, and DeepSeek-R1-8B decided better (0.79 vs.\ 0.68, $p = 0.0192$) but about 9.9 times more slowly. Mistral-7B computed 29\% of 100 financial calculations correctly against 100\% for Python, a live quote cut price error from 210\% to 0.00\%, and hybrid retrieval reached 76\% Hit@5 on 50 filing questions. Yet in a live case study, both agent answers passed validation while describing AMD sample data under NVIDIA's and Alphabet's names. A controlled upper-bound study on the FinRank benchmark explains why: when one figure in the shared evidence is corrupted, self-consistency stays confident (agreement 0.55 to 0.59 while accuracy falls from 0.50 to 0.00), and a shared-evidence check (the kind our validator performs) catches 35\% of the errors, whereas a check against the original passage catches 77\% ($p = 5.96\times10^{-8}$); for DeepSeek-R1-8B, 68\% vs.\ 90\%. Verification must be independent of the evidence the writer saw.
\end{abstract}

\section{Introduction}
\label{sec:intro}

\begin{figure*}[!t]
\centering
\includegraphics[width=\textwidth]{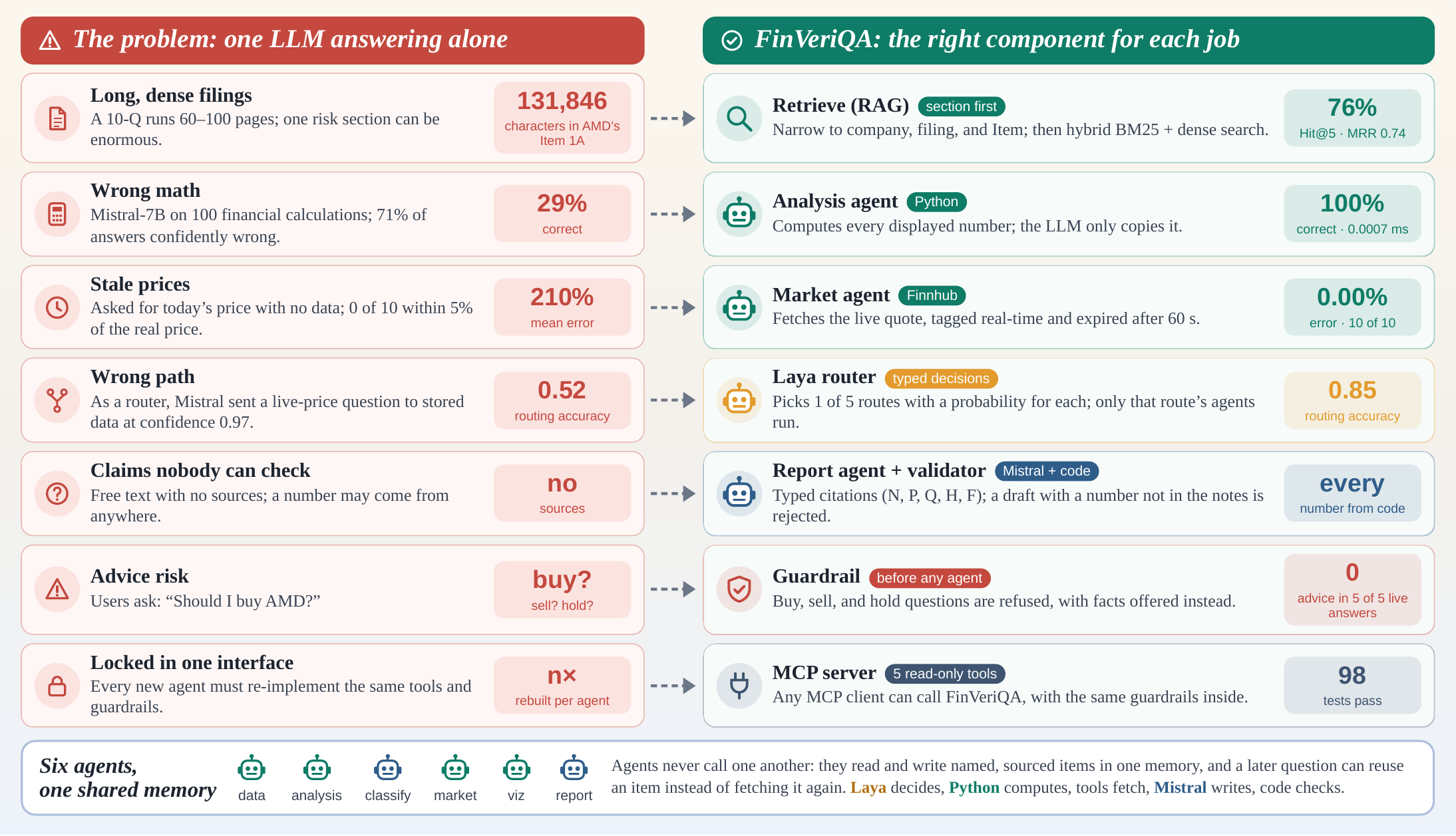}
\caption{The problem \sys{} addresses (left) and how it addresses it (right). The measured numbers come from our experiments and case study (\cref{sec:results}); each fix on the right is a separate component. The agents share evidence only through one memory (bottom), which also lets them reuse earlier work.}
\label{fig:problem}
\end{figure*}

Questions about public companies demand exact numbers and traceable sources. A quarterly report (Form 10-Q) can run 60 to 100 pages, and a reader who wants to know how revenue changed, which risks management disclosed, or where the stock trades today must combine structured figures, narrative text, and data that changes by the minute. Large language models (LLMs) answer such questions fluently, but they hallucinate \citep{ji2023hallucination}, and in finance a growth rate with the wrong sign is a substantive error, not a stylistic flaw. \Cref{fig:problem} summarizes the failures we measured when one LLM answers alone.

Prior work gives LLMs tools \citep{schick2023toolformer,yao2023react}, moves arithmetic into generated programs \citep{gao2023pal,chen2023pot}, and organizes models into agent teams \citep{wu2023autogen,hong2024metagpt,jin2025finsight}. Yet in most such systems the same generative model still makes the control decisions and states the numbers. We ask a narrower, measurable question: \emph{for each job inside a financial assistant (deciding the path a question takes, computing a number, retrieving evidence, and writing the answer), which component should do it?}

We answer it with \sys{}, an autonomous multi-agent research prototype for SEC filings built on open-source models. Six specialized agents each do one job: \emph{data} loads filings and figures, \emph{analysis} computes every number in Python, \emph{classify} labels passage sentiment, \emph{market} fetches the live quote, \emph{viz} draws charts, and \emph{report} writes the cited answer with Mistral-7B \citep{jiang2023mistral}. A compact typed-decision model, Laya \citep{laya}, routes each question so that only the agents it needs run, and screens out requests for investment advice. We evaluate a section-first hybrid retrieval-augmented generation (RAG) pipeline over SEC EDGAR filings \citep{secedgar} for filing evidence, and the market agent fetches live data from Finnhub \citep{finnhub}. Agents never call one another; they share one variable memory, adapted from \citet{jin2025finsight}, whose items later questions can reuse. A validator in code checks that every cited ID exists and every number appears in a computed note, every answer shows its route, confidence, steps, and sources, and a Model Context Protocol (MCP) server exposes the pipeline and its guardrails to other agents. \sys{} is autonomous in a specific sense: once a question arrives, it chooses its path, gathers and computes the evidence, writes, checks, and retries without a human in the loop. The prototype is incomplete in ways that matter here: its validator checks only that cited IDs exist and that numbers match the computed notes, its data agent still reads a documented sample file on two routes, and connecting retrieval and XBRL to the agents is in progress. We therefore measure the components one at a time (E1--E5, E7), together on live questions (E6), and the verification question in a controlled study (E8).

Measured live on a GPU cluster, the right component differed for every job. Laya routed better than Mistral (0.85 vs.\ 0.52), Mistral labeled sentiment better than Laya (0.93 vs.\ 0.88), and a per-task hybrid scored highest (0.88, an in-sample estimate). Scaling the LLM did not close the routing gap: Mistral-NeMo-12B decided no better than Mistral-7B (0.68 for both). A reasoning model, DeepSeek-R1-8B, did significantly better (0.77 and 0.79 in two runs; paired McNemar $p = 0.0192$) but was about 9.9 times slower per decision and still routed below Laya. Mistral-7B computed 29\% of financial calculations correctly against 100\% for Python, and gpt-5 reached 98\% on 50 problems. Missing data, not model weakness, explained invented prices (210\% error without a quote, 0.00\% with one). A live case study adds a caution: asked ``How did NVIDIA's revenue and margins change last quarter?'', \sys{} chose the right route at confidence 0.87, ran only the five steps that route needs, and passed every number through the validator, yet still answered with AMD sample figures under NVIDIA's name. A controlled upper-bound study on the FinRank benchmark \citep{mansouri2026finrank} shows the general mechanism: when agents share one wrong number, self-consistency stays confident and a check against the shared evidence passes the error, while verification against an independent source catches most of it (77\% against 35\%); the same holds for a reasoning model (90\% against 68\%). Our contributions are:
\begin{enumerate}
  \item \textbf{An open-source multi-agent architecture} that separates deciding, computing, retrieving, and writing: six specialized agents, a typed router, a shared memory with freshness tags and reuse, a code validator, explainable traces, and an MCP server with tested guardrails (\cref{sec:system}).
  \item \textbf{Controlled live experiments} on who should decide and who should compute: six decision strategies with bootstrap intervals and McNemar tests, an ablation of larger and reasoning LLMs as decision-makers, three LLMs on 100 financial calculations, and LLM prices with and without live data (\cref{sec:res-e1,sec:res-e7,sec:res-math,sec:res-live}).
  \item \textbf{A retrieval study} of four search modes over SEC filings and three embedding models (\cref{sec:res-rag}).
  \item \textbf{Evidence that shared-evidence verification is not enough:} a live case study with a failure taxonomy, and a controlled upper-bound study on the external FinRank benchmark in which independent provenance catches 77\% of corrupted-evidence errors against 35\% for a shared-evidence check, and 90\% against 68\% for a reasoning model (\cref{sec:res-e6,sec:res-e8}).
\end{enumerate}

\section{Related Work}
\label{sec:related}

\paragraph{Financial question answering and sentiment.} FinQA \citep{chen2021finqa}, TAT-QA \citep{zhu2021tatqa}, and ConvFinQA \citep{chen2022convfinqa} require numerical reasoning over tables and text; FinanceBench \citep{islam2023financebench} and DocFinQA \citep{reddy2024docfinqa} show that retrieval-augmented models still fail on real, full-length filings, and FinRank \citep{mansouri2026finrank} pairs filing questions with gold passages and curated hard negatives. BloombergGPT \citep{wu2023bloomberggpt} and FinGPT \citep{yang2023fingpt} adapt LLMs to finance, and FinBERT \citep{araci2019finbert} fine-tunes BERT on the Financial PhraseBank \citep{malo2014good} because general word lists misread financial text \citep{loughran2011liability}. Rather than training a financial model, we measure which existing component should perform each step, and we keep numbers out of the generator entirely.

\paragraph{Tools, programs, and agents.} Toolformer \citep{schick2023toolformer} and ReAct \citep{yao2023react} let a model call tools while it reasons; PAL \citep{gao2023pal} and Program-of-Thoughts \citep{chen2023pot} move arithmetic into code the model writes, avoiding many chain-of-thought errors \citep{wei2022cot}. Multi-agent frameworks coordinate agents through conversation \citep{wu2023autogen} or standard operating procedures \citep{hong2024metagpt}, echoing blackboard systems \citep{nii1986blackboard}, and the FinSight deep-research system of \citet{jin2025finsight} places data, tools, and agents in one variable space (CAVM). We adopt that shared memory for interactive question answering, add freshness tags and reuse, and go one step further than program-aided reasoning: the LLM writes neither the decision nor the program. In finance, a separate FinSight system \citep{srivastava2026finsight} coordinates multi-agent, multi-modal equity research, and FinSAgent \citep{chi2026finsagent} answers SEC-filing questions with five role-specialized agents tied to the 10-K Item structure, demoting similar but unreliable passages with a gate trained on hard negatives.

\paragraph{Retrieval.} RAG \citep{lewis2020rag} grounds answers in retrieved text. BM25 \citep{robertson2009bm25} matches terms, dense retrievers \citep{karpukhin2020dpr,reimers2019sbert} match meaning, reciprocal rank fusion \citep{cormack2009rrf} combines rankings, and cross-encoders re-rank candidates \citep{nogueira2019passage}. We use BGE embeddings \citep{xiao2023cpack} and evaluate Qwen3 embeddings \citep{zhang2025qwen3emb}; ALCE \citep{gao2023alce} and RAGAS \citep{es2024ragas} measure citation quality.

\paragraph{Routing, calibration, and protocols.} FrugalGPT \citep{chen2023frugalgpt} and RouteLLM \citep{ong2024routellm} route queries among LLMs; we route among tool paths with a non-generative model. Whether confidence can gate a decision depends on calibration \citep{guo2017calibration,naeini2015ece}, and LLM self-confidence is known to be imperfect \citep{kadavath2022know}; self-consistency \citep{wang2023selfconsistency} instead treats agreement among sampled answers as confidence, and we test when that agreement misleads. MCP \citep{anthropic2024mcp,mcpspec} standardizes how applications expose tools to model clients.

\paragraph{Verification and provenance.} Recent work checks agent outputs against their sources. VeriFin \citep{hall2026verifin} grounds the values a financial claim needs in filed XBRL facts, sets the required calculation independently of the candidate answer, and checks it with an SMT solver; FinGround \citep{guo2026finground} splits answers over financial documents into atomic claims, verifies each with a type-specific method that includes recomputing formulas, and rewrites unsupported claims with citations; ProvenanceGuard \citep{alvarez2026provenanceguard} routes each claim of an MCP-based agent's answer to source-specific evidence and flags claims attributed to the wrong source; and PARSE \citep{pai2026parse} sanitizes retrieved professional documents, including SEC filings, against prompt injection while checking that their facts are preserved. In multi-agent systems, unsupported claims can spread between agents and gain confidence through repetition, so agreement alone does not indicate correctness \citep{jamshidi2026collective}. We isolate a complementary failure: a wrong value inside a plausible passage passes both a check against the evidence the writer read and agreement among samples that read it, and is caught only by a check against an independent source.

\section{System Design}
\label{sec:system}

\begin{figure*}[!t]
\centering
\includegraphics[width=0.95\textwidth]{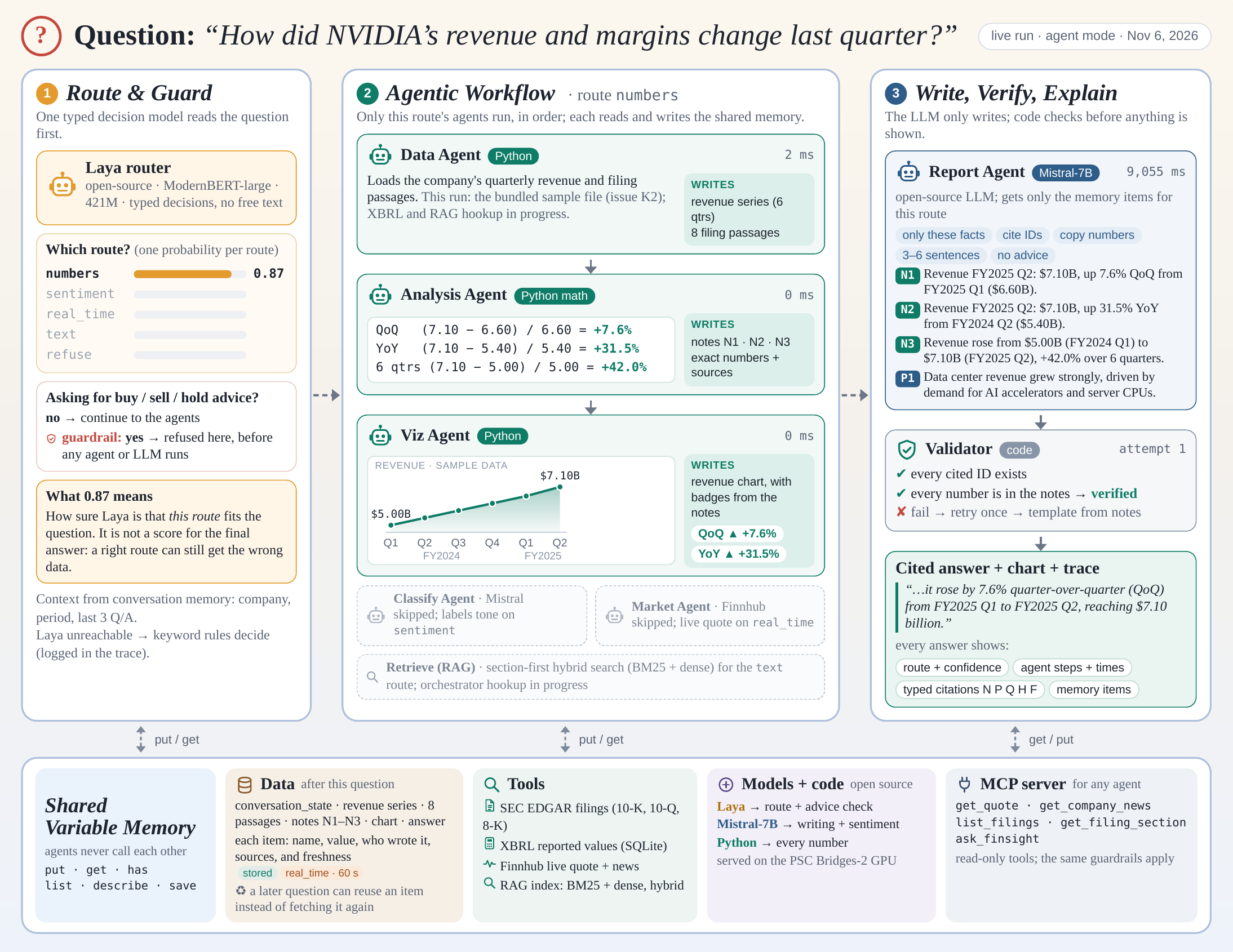}
\caption{Overview of \sys{} on one live question (case-study question Q4, \cref{sec:res-e6}). (1)~Laya routes the question and checks it for advice requests; its 0.87 is a confidence in the route, not in the answer. (2)~Only the \code{numbers} route's agents run: the data agent loads the figures, the analysis agent does the math in Python, and the viz agent draws the chart; the classify and market agents are skipped. (3)~Mistral-7B writes from the notes and passages, and a validator in code checks every cited ID and number before the answer, chart, and trace are shown. All agents read and write one shared variable memory (bottom), shown with the tools, the open-source models, and the MCP server. Times, values, and the quoted sentence are from the run; the figures came from the bundled AMD sample file (issue K2).}
\label{fig:overview}
\end{figure*}

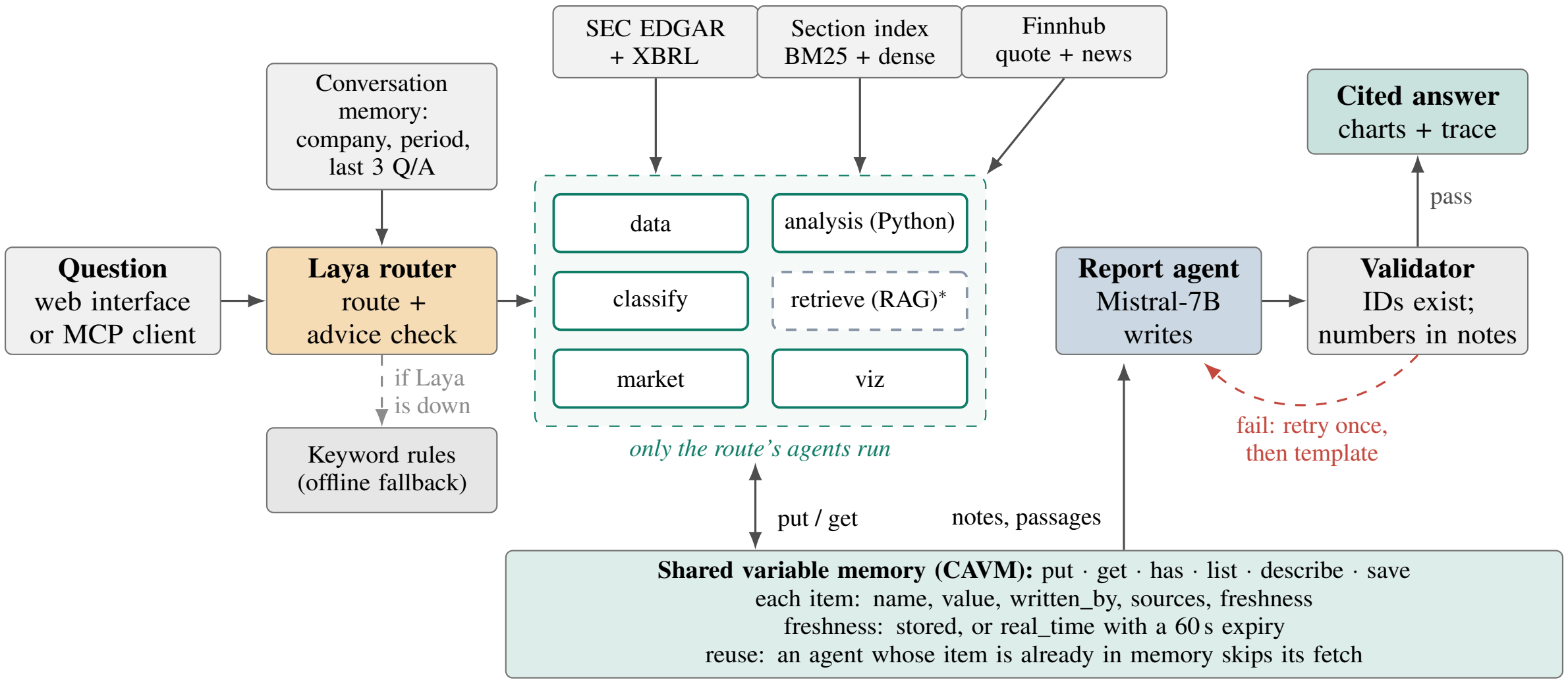
\begin{figure*}[!t]
\centering
\begin{tikzpicture}[
  font=\footnotesize,
  box/.style={draw=black!55, rounded corners=2pt, align=center, inner sep=3pt, minimum height=0.85cm},
  agent/.style={box, font=\scriptsize, fill=white, draw=pycol, line width=0.7pt, minimum width=1.95cm, minimum height=0.58cm},
  src/.style={box, font=\scriptsize, fill=black!5, text width=1.85cm, minimum height=0.72cm},
  arr/.style={-{Latex[length=1.8mm]}, black!70, line width=0.6pt},
]
\node[box, fill=black!6, text width=1.95cm] (q) at (1.05,0) {\textbf{Question}\\ web interface\\ or MCP client};
\node[box, fill=laycol!35, text width=2.1cm] (router) at (3.75,0) {\textbf{Laya router}\\ route +\\ advice check};
\node[box, fill=black!6, text width=2.1cm, font=\scriptsize] (conv) at (3.75,1.75) {Conversation memory:\\ company, period,\\ last 3 Q/A};
\node[box, fill=black!10, text width=2.1cm, font=\scriptsize] (rules) at (3.75,-1.7) {Keyword rules\\ (offline fallback)};
\node[agent] (data) at (6.45,0.78) {data};
\node[agent] (analysis) at (8.65,0.78) {analysis (Python)};
\node[agent] (classify) at (6.45,0) {classify};
\node[agent, dashed, draw=slatecol] (retrieve) at (8.65,0) {retrieve (RAG)$^{\ast}$};
\node[agent] (market) at (6.45,-0.78) {market};
\node[agent] (viz) at (8.65,-0.78) {viz};
\begin{scope}[on background layer]
\node[draw=pycol, dashed, rounded corners=3pt, fill=pycol!4, fit=(data)(analysis)(market)(viz), inner sep=5pt] (agents) {};
\end{scope}
\node[font=\scriptsize\itshape, text=pycol, anchor=north] at (agents.south) {only the route's agents run};
\node[src] (sec) at (6.5,2.6) {SEC EDGAR\\ + XBRL};
\node[src] (idx) at (8.55,2.6) {Section index\\ BM25 + dense};
\node[src] (fin) at (10.6,2.6) {Finnhub\\ quote + news};
\node[box, fill=miscol!25, text width=1.85cm] (report) at (11.55,0) {\textbf{Report agent}\\ Mistral-7B\\ writes};
\node[box, fill=black!8, text width=2.0cm] (val) at (14.15,0) {\textbf{Validator}\\ IDs exist;\\ numbers in notes};
\node[box, fill=pycol!22, text width=2.0cm] (ans) at (14.15,1.9) {\textbf{Cited answer}\\ charts + trace};
\node[box, fill=pycol!14, text width=10.4cm, font=\scriptsize, anchor=north] (mem) at (10.3,-2.5) {\textbf{Shared variable memory (CAVM):} put $\cdot$ get $\cdot$ has $\cdot$ list $\cdot$ describe $\cdot$ save\\ each item: name, value, written\_by, sources, freshness\\ freshness: stored, or real\_time with a 60\,s expiry\\ reuse: an agent whose item is already in memory skips its fetch};
\draw[arr] (q) -- (router);
\draw[arr] (conv) -- (router);
\draw[arr, dashed, black!45] (router) -- node[right, font=\scriptsize, align=left]{if Laya\\ is down} (rules);
\draw[arr] (router) -- (agents.west);
\draw[arr] (sec.south) -- (sec.south |- agents.north);
\draw[arr] (idx.south) -- (idx.south |- agents.north);
\draw[arr] (fin.south) -- (agents.north east);
\draw[{Latex[length=1.8mm]}-{Latex[length=1.8mm]}, black!70, line width=0.6pt] (7.5,-1.62) -- (mem.north -| 7.5,0);
\node[font=\scriptsize, anchor=west] at (7.6,-2.2) {put / get};
\draw[arr] (mem.north -| 11.2,0) -- (11.2,-0.62);
\node[font=\scriptsize, anchor=east] at (11.1,-2.2) {notes, passages};
\draw[arr] (report) -- (val);
\draw[arr] (val) -- node[right, font=\scriptsize]{pass} (ans);
\draw[arr, dashed, errcol] (val.south) to[bend left=45] node[below, font=\scriptsize, align=center, pos=0.5]{fail: retry once,\\ then template} (12.0,-0.62);
\end{tikzpicture}
\caption{\sys{} architecture. The six agents are data, analysis, classify, market, viz (green box), and report (blue). Laya routes each question; only the agents that route needs run; agents exchange evidence only through the shared memory, and may reuse an item already there; the report agent's Mistral writes from that evidence; and a validator in code checks every citation and number before the answer is shown. Keyword rules take over when Laya is unreachable. $^{\ast}$The retrieval (RAG) component supplies filing passages; its integration into the agent orchestrator is in progress.}
\label{fig:arch}
\end{figure*}

\Cref{fig:overview} follows one live question through \sys{}, and \cref{fig:arch} shows the full architecture.

\subsection{Design principles}
\label{sec:principles}
Six principles guide the design: (P1) numbers come from code or tools, never from the LLM, which may only copy them; (P2) structured decisions come from a decision model with usable confidence and a rule-based fallback; (P3) agents share evidence only through one memory, so every run can be inspected and replayed; (P4) every claim carries a typed citation; (P5) buy, sell, and hold questions are declined before any agent runs; and (P6) test doubles for both models let the full system run without a GPU.

\subsection{Data paths and shared memory}
\label{sec:memory}
Stored and real-time data follow separate paths. The \emph{batch path} holds filings (10-K, 10-Q, and earnings 8-Ks with the press release in Exhibit 99.1) from SEC EDGAR, exact XBRL values in SQLite, and a section index for retrieval (\cref{sec:rag}); the \emph{real-time path} fetches the current quote and company news from Finnhub. All agents read and write one shared memory through a small interface (\code{put}, \code{get}, \code{has}, \code{list}, \code{describe}, and \code{save}), and each item records its value, author, sources, and freshness (\code{stored}, or \code{real\_time} with a 60\,s expiry). Agents place the one-line \code{describe()} index in prompts instead of raw data, and \code{save()} writes a JSON snapshot that makes every answer traceable. A separate conversation memory keeps only the company, fiscal period, and last three question--answer pairs, which bounds prompt growth.

\paragraph{Memory reuse.} Because items describe themselves, an agent can check \code{has} before working and reuse an item already present; the trace then reads ``Reused from memory (skipped re-fetch)''. With the bundled sample file the saving is small (the data step takes about 2\,ms), but a live fetch is not (one filing-section call took 1{,}136\,ms, \cref{sec:res-mcp}), and fewer SEC EDGAR calls respect its fair-access limits. Expiry keeps reuse safe for prices. The case study (\cref{sec:res-e6}) shows the other side: reuse must be keyed to company, period, filing, and route, not to the item name alone.

\subsection{Agents and routing}
\label{sec:agents}
\label{sec:routing}
\Cref{tab:agents} lists the six agents. Only classify and report call the LLM, and neither computes numbers. The analysis agent writes each result as a note, for example ``Revenue FY2025 Q2: \$7.10B, up 7.6\% QoQ from FY2025 Q1 (\$6.60B)'', and the classify agent maps any reply outside its three labels to ``uncertain, needs review''. The data agent currently reads a documented sample file for the numbers and sentiment routes; connecting it to the XBRL store is in progress (issue K2).

\begin{table*}[!t]
\centering
\small
\caption{The six agents of \sys{}, and the retrieval component. Each agent reads from and writes to the shared memory; none calls another agent. $^{\ast}$Integration of the retrieval component into the orchestrator is in progress.}
\label{tab:agents}
\begin{tabularx}{\textwidth}{@{}l X X l@{}}
\toprule
\textbf{Agent} & \textbf{Reads} & \textbf{Writes} & \textbf{Engine} \\
\midrule
\code{data\_agent} & selected company & revenue series; filing passages & code \\
\code{analysis\_agent} & revenue series & notes N1--N3: QoQ, YoY, and trend, with exact numbers and sources & Python \\
\code{classify\_agent} & filing passages & one label per passage (positive, neutral, negative); sentiment breakdown & Mistral \\
\code{market\_agent} & selected company & live quote, tagged \code{real\_time} (60\,s expiry) & code + Finnhub \\
\code{viz\_agent} & revenue series; sentiment breakdown & chart specifications, rendered with Altair & code \\
\code{report\_agent} & notes, passages, quote, conversation state & the cited answer (3--6 sentences) & Mistral + validator \\
\midrule
retrieve (RAG)$^{\ast}$ & question, company, section & cited filing passages (\cref{sec:rag}) & retrieval \\
\bottomrule
\end{tabularx}
\end{table*}

Laya \citep{laya} is a non-autoregressive typed-decision model built on ModernBERT-large \citep{warner2024modernbert}, with about 421M parameters. Given the question and a list of typed questions, it returns a choice and a probability for every option; \sys{} asks two: which route, and whether the user wants investment advice. The five routes fix which agents run: \code{numbers} (data $\rightarrow$ analysis $\rightarrow$ viz $\rightarrow$ report), \code{sentiment} (data $\rightarrow$ classify $\rightarrow$ viz $\rightarrow$ report), \code{real\_time} (market $\rightarrow$ report), \code{text} (data or retrieval $\rightarrow$ report), and \code{refuse} (a polite refusal that offers facts). A wrong route yields a wrong answer even when every agent works, so routing matters: keyword rules route 0.65 of our routing questions correctly and refused ``share buybacks'' as advice until whole-word matching was added (issue K3), while Laya routes 0.85 (\cref{sec:res-e1}). If Laya is unreachable, keyword rules decide and the trace records the fallback.

\paragraph{What the confidence means.} The interface shows Laya's probability for the chosen route, for example ``Laya $\rightarrow$ numbers $\cdot$ confidence 0.87''. It measures how sure the router is that the path fits the question, not whether the answer will be right: in the case study two correct routes at 0.87 still produced answers about the wrong company, and in E3 a mean confidence of only 0.68 on price questions came with 10 of 10 correct routes. Confidence is useful for gating the route (for instance, asking a clarifying question below a threshold tuned on development data), provided it is calibrated, which our harness checks with expected calibration error.

\subsection{Writing, validation, guardrails, and explainability}
\label{sec:validator}
\label{sec:explain}
The report agent prompts Mistral with the notes, passages, and conversation state and five rules: use only these facts, cite IDs, copy numbers exactly, write three to six sentences, and give no advice. Each evidence type has a citation prefix: N (computed notes), P (passages), Q (quotes), H (headlines), and F (filing sections). Code then validates the draft (\cref{alg:report}). Because any draft with a number absent from the notes is rejected and the fallback is built from the notes, every number in a displayed answer comes from code. The validator does \emph{not} check that a note belongs to the company asked about, that a cited passage supports its sentence, or that every part of the question is answered.

\begin{algorithm}[!t]
\caption{Report generation with validation}
\label{alg:report}
\footnotesize
\begin{algorithmic}[1]
\Require notes $\mathcal{N}$ computed in Python, passages $\mathcal{P}$, conversation state $c$
\Function{Valid}{$d$}
  \State \Return $\textsc{CitedIDs}(d) \subseteq \textsc{IDs}(\mathcal{N} \cup \mathcal{P})$ \textbf{and} $\textsc{Numbers}(d) \subseteq \textsc{Numbers}(\mathcal{N})$
\EndFunction
\State $d \gets \textsc{Mistral}(\textsc{Prompt}(\mathcal{N}, \mathcal{P}, c))$ \Comment{five rules}
\If{\Call{Valid}{$d$}} \Return $d$ \EndIf
\State $d \gets \textsc{Mistral}(\textsc{Prompt}(\mathcal{N}, \mathcal{P}, c))$ \Comment{one retry}
\If{\Call{Valid}{$d$}} \Return $d$ \EndIf
\State \Return $\textsc{Template}(\mathcal{N})$ \Comment{built from the notes}
\end{algorithmic}
\end{algorithm}

\paragraph{Guardrails and explainability.} Guardrails are enforced in code, not by prompt: advice requests are refused before any agent or LLM runs; unsupported numbers and missing citations are rejected; real-time values expire; each answer reports its data mode, so sample data is flagged; and through MCP every tool is read-only, news is labeled as untrusted text, API keys never reach the logs, and a failed quote returns an error rather than sample data. Every answer also explains itself at four levels: \emph{why this path} (route and confidence), \emph{what ran} (each agent, its engine, and its time, with skipped and reused steps marked), \emph{where each claim comes from} (typed citations with the cited text), and \emph{what it was built from} (the memory listing and the validator's verdict). This detail is what exposed every failure in our case study.

\subsection{Retrieval over filing text}
\label{sec:rag}
Retrieval serves questions about what a filing \emph{says}; numbers come from XBRL and live values from Finnhub. Each filing is split by Item (e.g., Item 1A Risk Factors, Item 7 MD\&A) and then into passages of about 220 words (at most 300; passages under 30 words are merged with neighbors); tables are excluded. Each passage receives a dense \code{bge-small-en-v1.5} embedding \citep{xiao2023cpack} (384 dimensions, computed with FastEmbed \citep{fastembed}) and BM25 weights \citep{robertson2009bm25}, and is stored in Qdrant \citep{qdrant} with its ticker, form, accession number, section, and link. At query time the retriever filters to the company, latest filing, and section. A section that fits a 12{,}000-token budget is passed whole; otherwise it is searched in one of four modes: sparse, dense, hybrid with reciprocal rank fusion \citep{cormack2009rrf} (the default), or hybrid re-ranked over the top 30 by a MiniLM cross-encoder \citep{nogueira2019passage,wang2020minilm,bajaj2016msmarco}. Duplicate passages are removed, keeping the newest. Section extraction itself needed care: every 10-K Item heading appears twice, in the table of contents and in the body, and our first extractor returned a 40-character stub for NVIDIA's Item 1A; choosing the occurrence followed by the most text recovered 114{,}916 characters.

\subsection{Interfaces and infrastructure}
\label{sec:mcp}
A web interface built with Streamlit \citep{streamlit} has two modes: a \emph{filing-QA mode}, which extracts a filing's main Items and lets Mistral answer with section citations (without the router or agents), and an \emph{agent mode}, which runs the routed pipeline and shows the trace, Altair charts \citep{vanderplas2018altair}, and memory. An MCP server (FastMCP over stdio) wraps the same code in five read-only tools that return JSON with citation IDs: \code{get\_quote}, \code{get\_company\_news}, \code{list\_filings}, \code{get\_filing\_section}, and \code{ask\_finsight}. Both models run on a Tesla V100 on PSC Bridges-2 \citep{brown2021bridges2}: prebuilt vLLM \citep{kwon2023vllm} wheels lack kernels for this GPU, so Mistral-7B-Instruct is served by Ollama \citep{ollama} behind an OpenAI-compatible endpoint and Laya by a FastAPI server, both reached through SSH tunnels.

\section{Evaluation Methodology}
\label{sec:method}

We ask eight questions: who should make structured decisions (E1, E3); whether a larger or a reasoning LLM would make them better (E7); whether an LLM can compute the numbers an assistant displays, and whether newer models change that (E2a, E4, E5); whether wrong prices come from the model or from missing data (E2b); which retrieval configuration finds the right evidence (R1, R2); whether the pipeline can be exposed through MCP without weakening its guardrails (M1); what fails when all components run together (E6); and whether verification survives evidence errors that every agent shares (E8).

\paragraph{Data.} \textbf{D1} (E1, E7): 100 labeled items: 60 routing questions (numbers 15, text 12, real\_time 10, sentiment 10, refuse 8, out of scope 5; 26 follow-ups) and 40 filing sentences on NVIDIA and AMD (14 positive, 13 neutral, 13 negative), with model-assisted gold labels pending full human sign-off. Each routing question also carries an advice decision, so advice recall is computed over the 8 \code{refuse} questions and the false-refusal rate over the other 52. The harness can split each set 50/50 into development and test halves, but because the set is small, E1 and E7 report all 100 items. \textbf{D2} (E2a, E4, E5): 100 seeded financial-math problems of eight types with exact Python answers; E4 uses the first 50. \textbf{D3} (E2b): ten large-cap tickers scored against the live Finnhub quote. \textbf{D4} (R1): 50 filing questions; a passage is correct when it matches the expected content pattern. \textbf{D5} (R2): six questions over NVIDIA and Apple filings and a fixed 144-passage corpus. \textbf{D6} (E6): five questions asked through the web interface on November 6, 2026, logged verbatim with their traces and judged by the authors against the sources shown. \textbf{D7} (E8): FinRank \citep{mansouri2026finrank}, 1{,}185 manually authored question--answer records over 10-K and 10-Q filings of 22 companies in pharmaceuticals, oil and gas, and automotive (filing years 2024--2025), each with a reference answer, gold supporting passages, and curated hard negatives; we use the 413 records whose reference-answer figure appears verbatim in a gold passage and sample 60.

\paragraph{Decision strategies (E1).} Six setups make the same routing and sentiment decisions, with Mistral writing in all of them: \textbf{A}~keyword rules; \textbf{B}~Mistral alone; \textbf{C}~Laya alone; \textbf{D}~a cascade (Laya, then Mistral below a confidence threshold); \textbf{E}~an agreement rule (disagreements marked for review); and \textbf{F}~a per-task hybrid (Laya routes, Mistral labels sentiment). Mistral's confidence is the probability of its label token; Laya's is the predicted option's probability. The cascade uses a fixed confidence threshold.

\paragraph{LLM ablation (E7).} Because Mistral-7B routed poorly, we asked whether a larger model of the same family, or a model that reasons before answering, would close the gap. E7 repeats setup~B (the LLM alone as decision-maker, with no keyword rules and no Laya), which is the only setup whose behavior depends on the choice of LLM. We compare Mistral-7B with two alternatives: Mistral-NeMo-12B, a 12B model that Mistral AI built with NVIDIA \citep{mistral2024nemo}, as a size control; and DeepSeek-R1-8B, an 8B distillation of the DeepSeek-R1-0528 update \citep{deepseek2025r1}, which writes a chain of thought inside a \code{<think>} block before its answer. All three are Ollama's default 4-bit quantizations (\code{mistral:latest}, digest \code{6577803aa9a0}, 4.4\,GB; \code{mistral-nemo:latest}, digest \code{e7e06d107c6c}, 7.1\,GB; \code{deepseek-r1:8b}, digest \code{6995872bfe4c}, 5.2\,GB). Task, data, and prompt are identical. For the two Mistral models, each decision is a forced one-word classification (``Answer with exactly one word from: \{options\}''), decoded greedily (temperature 0, at most 5 tokens) with log-probabilities; a decision's confidence is the probability of its first generated token, and an output is \emph{valid} when its first word parses to an allowed label. This protocol cannot score a reasoning model, whose first tokens are reasoning text, so R1 is given up to 1{,}024 tokens at temperature 0.6, DeepSeek's recommended setting; the \code{<think>} block is removed and the final label is parsed from the remaining text. All three models are scored on the same gold data with the same metric code, and the Mistral-7B row, re-measured through the harness used for R1, reproduced the original run to four decimals. Because R1 samples at temperature 0.6, we ran it twice; the second run saved per-item results, and we use it for an exact McNemar test of R1 against Mistral-7B on the 100 intent and sentiment items. We report accuracy over the 100 intent and sentiment items with a 1{,}000-resample bootstrap 95\% CI, intent accuracy and macro-F1 over the six intent classes, sentiment accuracy and macro-F1 over the three sentiment classes, advice recall over the 8 advice requests and the false-refusal rate over the other 52 routing questions (each routing question carries one advice decision), the valid-output rate, and mean latency per decision.

\paragraph{Verification under shared evidence errors (E8).} We test the hypothesis that self-consistency produces false confidence when agents share the same erroneous evidence, and that independently grounded, provenance-aware verification reduces such correlated errors. For each FinRank item we choose a target figure that appears in both the reference answer and a gold passage, and build two contexts from a window of that passage: \emph{clean}, the passage unchanged, and \emph{corrupted}, the same passage with the target figure replaced by a plausible wrong value (the true value multiplied by 1.18, 1.27, 0.82, or 0.73). The corrupted passage is on-topic and well formed; only its number is wrong, as when every agent reads one bad copy of the evidence. Mistral-7B (\code{mistral:latest}, 4-bit, on the Jetstream2 A100 of E7) answers under three conditions, all scored by number grounding rather than free-text judging. (A)~\emph{Self-consistency}: five samples at temperature 0.7; the model's figure is the one in the context closest to its answer by significant figures, confidence is the majority agreement, and accuracy is whether the majority figure equals the true target. (B)~\emph{Shared-evidence check}, the kind of check \sys{}'s validator performs: an answer passes if its figure appears in the retrieved context. (C)~\emph{Independent provenance}, the proposed check: an answer passes only if its figure appears in the original, uncorrupted gold passage. Because C consults the gold passage, it is an oracle reference, and E8 measures an \emph{upper bound} on what independent provenance can catch rather than a deployed check. A \emph{silent error} is an answer that differs from the true target and is not flagged. We report each check's detection rate on corrupted evidence with bootstrap 95\% intervals, an exact McNemar test of C against B, and the silent-error rate as the fraction $p$ of corrupted evidence grows from 0 to 1. To test whether reasoning resists value corruption, we repeated the experiment with DeepSeek-R1-8B (\code{deepseek-r1:8b}, as in E7) on 40 items.

\paragraph{Metrics and setup.} A unit-tested NumPy harness computes accuracy, macro-F1, expected calibration error \citep{naeini2015ece,guo2017calibration}, Hit@5, Recall@5, and mean reciprocal rank (MRR), with 95\% bootstrap intervals \citep{efron1979bootstrap} and McNemar tests \citep{mcnemar1947,dietterich1998tests}. An LLM's calculation counts as correct within 0.1 absolute or 1\% relative of the Python value; any other number is \emph{confidently wrong}. Each model was asked to reply with only the number, and the last number in its reply was parsed. E1--E3 and E6 ran live on Bridges-2 (Mistral-7B-Instruct via Ollama on a V100; Laya over HTTP). E4 called gpt-5 \citep{openai2025gpt5} with minimal reasoning effort, and E5 ran Ministral~3~8B \citep{liu2026ministral} with Ollama on a Colab T4 with a 4{,}096-token output limit. E7 served all three models with Ollama's OpenAI-compatible endpoint from one NVIDIA A100-SXM4-40GB on Jetstream2 \citep{hancock2021jetstream2}, reached through an SSH tunnel. Latencies therefore come from different hardware and are indicative.

\section{Results}
\label{sec:results}

\subsection{Who should decide? (E1)}
\label{sec:res-e1}

\begin{table}[!t]
\centering
\small
\caption{E1: decision strategies on all 100 items of D1. Overall accuracy with 95\% bootstrap CI; routing and sentiment accuracy separately. Mistral writes the answer in every setup.}
\label{tab:e1}
\begin{tabular}{@{}lccc@{}}
\toprule
\textbf{Setup} & \textbf{Overall (95\% CI)} & \textbf{Routing} & \textbf{Sentiment} \\
\midrule
A rules   & 0.62 [0.53, 0.71] & 0.65 & 0.57 \\
B Mistral & 0.68 [0.60, 0.77] & 0.52 & \textbf{0.93} \\
C Laya    & 0.86 [0.79, 0.92] & \textbf{0.85} & 0.88 \\
D cascade & 0.74 [0.66, 0.83] & 0.62 & \textbf{0.93} \\
E agree   & 0.65 [0.57, 0.75] & 0.52 & 0.85 \\
\rowcolor{pycol!12} F hybrid  & \textbf{0.88 [0.82, 0.94]} & \textbf{0.85} & \textbf{0.93} \\
\bottomrule
\end{tabular}
\end{table}

\Cref{tab:e1} shows that no single model is best at both decisions. \textbf{Routing belongs to Laya:} it routed 0.85 of the 60 routing questions correctly (51), against 0.52 for Mistral (31) and 0.65 for keyword rules; over all 100 decisions, Laya alone was right on 86 and Mistral alone on 68, a significant difference (McNemar $p = 0.0003$; Laya right and Mistral wrong on 21 items, the reverse on 3). Mistral was confidently wrong, sending ``What is AMD's stock price right now?'' to the numbers route with probability 0.97. \textbf{Sentiment belongs to Mistral:} 0.93, ahead of Laya (0.88) and rules (0.57), which read ``free cash flow increased significantly'' as negative, a known weakness of word lists \citep{loughran2011liability}. \textbf{The per-task hybrid scored highest:} setup F reached 0.88 overall (95\% CI [0.82, 0.94]); its interval excludes rules, Mistral alone, and the agreement rule, but overlaps Laya alone (0.86) and the cascade (0.74), so we do not claim it is significantly better than those two. Because F combines the component that did best on each task on these same 100 items, its 0.88 is an in-sample estimate. \textbf{A weaker fallback hurts:} the cascade routed 0.62 and the agreement rule 0.52, because the fallback (Mistral) was a weaker router than the primary (Laya).

\subsection{Does a larger or a reasoning LLM decide better? (E7)}
\label{sec:res-e7}

\begin{table}[!t]
\centering
\footnotesize
\setlength{\tabcolsep}{3.5pt}
\caption{E7: three LLMs as the sole decision-maker (setup B) on all items of D1, with the same gold data and metric code. Accuracy covers the 60 intent and 40 sentiment items; advice recall covers the 8 advice requests and the false-refusal rate the other 52 routing questions. Values are shown at the precision measured; ``---'' means not reported. The upper block is R1's first run, with all metrics; the lower block is R1's second run, which saved per-item results and is used for the paired McNemar test ($b$: R1 right and Mistral-7B wrong; $c$: the reverse).}
\label{tab:e7}
\begin{tabular}{@{}lccc@{}}
\toprule
 & \textbf{Mistral-7B} & \textbf{NeMo-12B} & \textbf{R1-8B} \\
\midrule
Accuracy & 0.6800 & 0.68 & \textbf{0.7700} \\
\quad 95\% CI & [0.60, 0.77] & [0.60, 0.77] & [0.69, 0.85] \\
Intent accuracy & 0.5167 & --- & \textbf{0.6667} \\
Intent macro-F1 & 0.3683 & 0.35 & \textbf{0.4784} \\
Sentiment accuracy & 0.9250 & 0.90 & 0.9250 \\
Sentiment macro-F1 & 0.9212 & --- & 0.9244 \\
Advice recall & 0.5000 & 0.38 & 0.5000 \\
False-refusal rate & 0.0769 & 0.00 & 0.0192 \\
Valid-output rate & 0.9700 & 0.99 & 1.0000 \\
Latency per decision & 267$^{a}$ / 262$^{b}$\,ms & 305\,ms & 2{,}588\,ms \\
Decoding & greedy, 5 tok. & greedy, 5 tok. & $T{=}0.6$, 1{,}024 tok. \\
Size on disk (4-bit) & 4.4\,GB & 7.1\,GB & 5.2\,GB \\
\midrule
\multicolumn{4}{@{}l}{\emph{R1 second run (per-item results)}} \\
Accuracy & 0.68 & --- & \textbf{0.79} \\
McNemar vs.\ 7B & & & $b{=}15$, $c{=}4$ \\
\quad exact $p$ & & & \textbf{0.0192} \\
\bottomrule
\end{tabular}

\smallskip
{\scriptsize $^{a}$Run alongside NeMo-12B.\quad $^{b}$Re-run alongside R1 through the same harness.}
\end{table}

\paragraph{Size: no gain.} Scaling the decision-maker from 7B to 12B did not help (\cref{tab:e7}). Overall accuracy was identical, 0.68 for both models, with fully overlapping 95\% bootstrap intervals ([0.60, 0.77] for each), so there is no evidence that the larger model decides better. On the two sub-tasks the 12B model was marginally worse: intent macro-F1 0.35 against 0.37, and sentiment accuracy 0.90 against 0.93. (From the reported totals, intent accuracy is 31 of 60 questions for the 7B model, the 0.52 of E1, and 32 of 60 for the 12B model; the larger model answered one more routing question correctly but with a lower macro-F1.) The models traded errors on the advice gate: the 12B model never refused a harmless question (false-refusal rate 0.00 against 0.08, that is, 0 against 4 of 52) but caught fewer real advice requests (recall 0.38 against 0.50, that is, 3 against 4 of 8), because it answered ``no'' more often overall. Both almost always produced a valid one-word label (0.97 and 0.99). The 12B model was about 14\% slower per decision (305 against 267\,ms) and 61\% larger on disk (7.1 against 4.4\,GB). The Mistral-7B results also reproduce setup B of E1 exactly (0.68 [0.60, 0.77]; sentiment 0.93) on different hardware. Because intent macro-F1 stays near 0.36 at both sizes, the bottleneck appears to be Mistral's fit for forced one-word intent classification rather than model capacity.

\paragraph{Reasoning: a gain on routing, at a cost in latency.} DeepSeek-R1-8B raised overall accuracy from 0.68 to 0.77, that is, from 68 to 77 of the 100 items (95\% CI [0.69, 0.85] against [0.60, 0.77]). Because R1 samples at temperature 0.6, its score varies between runs: a second run, the one that saved per-item results, scored 79 of 100 against the same 68 for Mistral-7B. On that matched run the gain is significant: R1 was right and Mistral-7B wrong on 15 items, and the reverse held on 4 (19 discordant pairs; exact McNemar $p = 0.0192$). Accuracy and the paired test are therefore reported from the same run, and the first run's 0.77 serves as a replicate. In the first run, all nine extra correct items came from routing intent, where the baseline was weakest: intent accuracy rose from 0.5167 to 0.6667 (31 to 40 of 60 questions), and intent macro-F1 from 0.3683 to 0.4784 (+0.110). Sentiment did not change (accuracy 0.9250 for both; macro-F1 0.9212 and 0.9244); at 0.90--0.93 for all three models, this sub-task has little headroom. On the advice gate, R1 matched Mistral-7B's recall (0.5000, 4 of 8) with fewer false refusals (0.0192 against 0.0769, that is, 1 against 4 of 52), and every one of its outputs parsed to a valid label (1.0000). Size alone does not explain the gain: the 12B Mistral, larger than R1 on disk (7.1 against 5.2\,GB), gained nothing, while R1, somewhat larger than Mistral-7B (5.2 against 4.4\,GB), gained nine items. The cost is latency: 2{,}588 against 262\,ms per decision in the same run, about 9.9 times slower (414 against 42\,s of wall time for the same 160 model calls). And R1 still trails Laya, which routed 0.85 of the same 60 questions in E1 at about 0.3\,s per question in E3. We therefore keep Mistral-7B as the writer and sentiment labeler, route with Laya, and keep R1 off the real-time routing path; no part of the configuration was changed as a result of E7.

\subsection{Can an LLM compute the numbers? (E2a, E4, E5)}
\label{sec:res-math}

\begin{table}[!t]
\centering
\footnotesize
\setlength{\tabcolsep}{4pt}
\caption{E2a, E4, E5: the same financial-math problems, four ways. Accuracy with the tolerance of \cref{sec:method}; every engine ran all 100 problems except gpt-5 (first 50). Times come from different hardware and are indicative.}
\label{tab:e45}
\begin{tabular}{@{}lccl@{}}
\toprule
\textbf{Engine} & \textbf{Accuracy} & \textbf{Time / calc} & \textbf{Hardware} \\
\midrule
\rowcolor{pycol!12} Python & \textbf{100\%} & 0.0007\,ms & CPU \\
gpt-5 (minimal effort) & 98\% (49/50) & 1{,}323\,ms & OpenAI API \\
Ministral 3 8B & 72\% (74\% on 50) & 0.3\,s & Colab T4 \\
Mistral-7B-Instruct & 29\% & 386\,ms & V100 (PSC) \\
\bottomrule
\end{tabular}
\end{table}

Mistral-7B answered every problem with a number and was right on 29\% (\cref{tab:e45}); 71\% of its answers were confidently wrong. It scored 0\% on the metrics an assistant reports most (quarter-over-quarter change, year-over-year change, and gross margin), and 17\% on three-quarter averages, 23\% on net margin, 58\% on absolute change, 67\% on percent of total, and 75\% on EPS. Its errors were plausible: a decline from \$25.64B to \$23.22B ($-9.44\%$) became $+6.83\%$, a year-over-year change of $-30.42\%$ became $+31.48\%$, a 42.63\% gross margin became 23.93\%, and an EPS of 3.13 became 0.32. Newer models narrow the gap but do not close it. Ministral~3~8B reached 72\% (74\% on the first 50), with margins and EPS now correct, but it dropped the minus sign on every wrong change answer ($-11.30$ became $11.30$), and its multi-step percentage changes stayed below 50\%. gpt-5 solved 49 of 50, missing one three-quarter average (24.71 instead of 24.05), using 3{,}921 tokens for the 50 problems. Python was exact and, per calculation, about 540{,}000 times faster than Mistral and 1.9 million times faster than gpt-5, which is why every displayed number in \sys{} is computed in code (P1).

\subsection{Live data and routing to tools (E2b, E3)}
\label{sec:res-live}

\begin{table}[!t]
\centering
\small
\caption{E2b: Mistral asked for current prices, without and with the live Finnhub quote. Top: summary over 10 tickers. Bottom: six of the ten answers without data.}
\label{tab:e2b}
\begin{tabular}{@{}lcc@{}}
\toprule
\textbf{Condition} & \textbf{Mean error} & \textbf{Within 5\%} \\
\midrule
Without data & 210\% & 0 / 10 \\
With live quote & \textbf{0.00\%} & \textbf{10 / 10} \\
\bottomrule
\end{tabular}

\vspace{6pt}
\begin{tabular}{@{}lrrr@{}}
\toprule
\textbf{Ticker} & \textbf{Live price} & \textbf{Mistral alone} & \textbf{Error} \\
\midrule
NVDA & \$217.95 & \$163.00 & 25\% \\
AMD & \$559.31 & \$103.00 & 82\% \\
AAPL & \$346.96 & \$134.70 & 61\% \\
TSLA & \$407.43 & \$1.00 & 100\% \\
GOOGL & \$322.88 & \$1{,}630.00 & 405\% \\
AMZN & \$254.21 & \$3{,}400.00 & 1{,}237\% \\
\bottomrule
\end{tabular}
\end{table}

Without data, Mistral's prices were off by 210\% on average and none of the ten was within 5\% of the live price (\cref{tab:e2b}); the GOOGL and AMZN answers match prices from before those stocks split, a stale-memory hallucination. Given the live quote, the same model reproduced every price exactly (0.00\% error, 10 of 10). The cause was missing data, and the fix is a data path, not a larger model. Laya then sends each question to the component that produces the right value: on the same inputs it routed all 10 price questions to \code{real\_time} (mean confidence 0.68, 319\,ms per question) and all 100 calculation questions to \code{numbers} (0.97, 315\,ms). These clean, single-intent questions are easier than E1's mixed set, on which Laya routed 0.85.

\subsection{Retrieval: search modes and embedders (R1, R2)}
\label{sec:res-rag}

\begin{table*}[!t]
\centering
\small
\begin{minipage}[t]{0.42\textwidth}
\centering
\caption{R1: search modes on 50 filing questions (D4).}
\label{tab:r1}
\begin{tabular}{@{}lcc@{}}
\toprule
\textbf{Mode} & \textbf{Hit@5} & \textbf{MRR} \\
\midrule
Sparse (BM25) & 74\% & 0.70 \\
Dense (bge-small) & 76\% & 0.73 \\
\rowcolor{pycol!12} Hybrid (default) & \textbf{76\%} & \textbf{0.74} \\
Rerank (cross-encoder) & 76\% & 0.74 \\
\bottomrule
\end{tabular}
\end{minipage}\hfill
\begin{minipage}[t]{0.54\textwidth}
\centering
\caption{R2: embedder pilot, dense search on a fixed 144-passage corpus (D5).}
\label{tab:r2}
\begin{tabular}{@{}lccc@{}}
\toprule
\textbf{Embedder} & \textbf{Hit@5} & \textbf{Recall@5} & \textbf{MRR} \\
\midrule
bge-small-en-v1.5 (current) & 0.833 & 0.500 & 0.652 \\
\rowcolor{pycol!12} Qwen3-Embedding-0.6B & \textbf{1.000} & \textbf{0.717} & \textbf{0.875} \\
Qwen3-Embedding-4B & 0.833 & 0.633 & 0.842 \\
\bottomrule
\end{tabular}
\end{minipage}
\end{table*}

Hybrid search found a correct passage in the top five for 76\% of the 50 questions, with MRR 0.74 (\cref{tab:r1}), up from 56\% (MRR 0.49) in an earlier configuration, partly because the evaluation now searches the latest 10-K. Dense search matched hybrid on Hit@5 with a lower MRR, sparse search trailed slightly, and re-ranking tied, so hybrid remains the default. For ``How do export limits affect Nvidia?'', hybrid search returned five on-topic passages out of five, against four for each of the other modes. In the embedder pilot (\cref{tab:r2}), Qwen3-Embedding-0.6B \citep{zhang2025qwen3emb} found a correct passage for every question and raised MRR from 0.652 to 0.875 (+34\%) over bge-small; the 4B model, about seven times larger, scored lower on all three metrics. Because the pilot is small and lenient, it nominates Qwen3-Embedding-0.6B for the full hybrid evaluation rather than settling the choice.

\subsection{MCP server and testing (M1)}
\label{sec:res-mcp}
The MCP test suite passed in full in continuous integration: 98 tests in 11.56\,s (82 server, 16 agent-core) across direct calls, an in-memory MCP session, and a real stdio subprocess, with a test for each guardrail (advice refused before any agent runs, keys never logged, stdout carrying only JSON-RPC, memory cleared on a ticker change, quote failures returned as errors, news labeled untrusted). In one offline session with the MCP Inspector, \code{get\_quote} took 111\,ms, \code{list\_filings} 69\,ms, \code{get\_company\_news} 345\,ms, \code{get\_filing\_section} 1{,}136\,ms, and \code{ask\_finsight} 48--58\,ms with the fake LLM; a general-purpose coding agent (Cursor) chose the tools on its own, quoted a live NVIDIA price with its citation, and relayed \sys{}'s refusal of ``Should I buy AMD?''. Testing found seven issues, five now fixed: ticker-blind memory reuse (K1), substring matching that refused ``share buybacks'' (K3), refusals tagged as sample data (K10), a wrong step label (K11), and one citation prefix with two meanings (D3). The bundled sample data (K2) and an indefinitely cached ``latest'' filing (K9) remain open.

\subsection{Five questions answered live (E6)}
\label{sec:res-e6}
E1--E5 test one component at a time. To see the components together, we asked five questions through the web interface with both models served from the GPU: three in filing-QA mode (no router) and two in agent mode (routed agents). \Cref{tab:e6} summarizes the answers and \cref{tab:e6-errors} groups the failures.

\begin{table*}[!t]
\centering
\footnotesize
\caption{E6: five questions answered live on November 6, 2026 (Mistral-7B and Laya on the Bridges-2 GPU). Q1--Q3: filing-QA mode (no router). Q4--Q5: agent mode (routed agents). All five answers gave no investment advice.}
\label{tab:e6}
\begin{tabularx}{\textwidth}{@{}L{0.45cm} L{2.95cm} L{3.0cm} Y Y@{}}
\toprule
 & \textbf{Question} & \textbf{Route and steps} & \textbf{What worked} & \textbf{What went wrong} \\
\midrule
Q1 & ``What are the main risk factors?'' (GOOGL) & no router; cites F2, F3, F4 & points 1--6 are genuine risk factors (advertising dependence, privacy changes, ad blocking, competition, the economy, new-business bets), each cited & points 7--8, cited to F4, are a pending acquisition (Wiz, Intersect) and a \$2.1B compensation charge: facts, not risks \\
\addlinespace
Q2 & ``Summarize management's discussion of results.'' (META) & no router; cites F1--F4 & the one MD\&A paragraph (F3) gives 2025 total revenue, \$200.97B & 3 of 4 paragraphs summarize other Items (business, risk summary, market risk); the figure is copied from text and not checked against XBRL \\
\addlinespace
Q3 & ``What is the tone of this filing?'' (AMD 10-K) & no router; cites F1--F4, all at the end & a fair overall read (``cautious and informative''); all four Items extracted in full (Item 1A: 131{,}846 characters) & citations cannot be matched to claims; no per-passage labels; a filler last sentence \\
\addlinespace
Q4 & ``How did NVIDIA's revenue and margins change last quarter?'' & Laya $\rightarrow$ \code{numbers}, 0.87; data, analysis, viz, report ran; classify, market skipped & every number matches a Python note (\$7.10B; $+$7.6\% QoQ; $+$31.5\% YoY; $+$42.0\% over six quarters); verified on attempt 1 & AMD sample data presented as NVIDIA's; margins never answered; P2 cited for a claim P2 does not contain \\
\addlinespace
Q5 & ``What's the tone and outlook in Alphabet's latest 10-K?'' & Laya $\rightarrow$ \code{sentiment}, 0.87; data reused from memory; classify, viz, report ran; analysis, market skipped & 8 passages labeled (5 positive, 3 negative), 7 of them clearly right; verified on attempt 1 & AMD sample data presented as Alphabet's; answer repeats revenue notes N1--N3 instead of the 5/3 split; same P2 mismatch; sentiment chart drew no bar \\
\bottomrule
\end{tabularx}
\end{table*}

\paragraph{What worked.} Both agent questions reached the expected route at confidence 0.87, and only that route's agents ran (\cref{fig:overview}). Every number in the NVIDIA answer (\$7.10B, 7.6\%, 31.5\%, \$5.40B, 42.0\%, \$5.00B) appears in Python notes N1--N3, and we recomputed all three changes from the quarterly values ($0.50/6.60$, $1.70/5.40$, $2.10/5.00$). The validator accepted both agent answers on the first attempt, the eight sentiment labels in Q5 were plausible (seven clearly right), and none of the five answers gave advice.

\paragraph{Verified did not mean correct.} Both agent answers described the bundled AMD sample file (K2) under the company asked about (``NVIDIA's revenue increased significantly'', ``the tone and outlook in Alphabet's latest 10-K''), with data-center, gaming, and console passages that fit AMD, not Alphabet; only the chart said SAMPLE DATA. Q4 never addressed margins, because the analysis agent computes only revenue changes. Both answers also cited P2 for a sentence that P2 does not contain. The sentences match the sixth and seventh sample passages, so the content is grounded but the identifier is wrong, which suggests that the prompt and the Sources list number passages differently. The validator passed all of this because it checks \emph{provenance} (each number came from a note and each cited ID exists), not \emph{relevance}, \emph{support}, or \emph{coverage}.

\paragraph{Context volume and memory reuse.} Given four Items at once, the filing-QA mode summarized what it was given rather than what was asked: in Q1, 2 of 8 points were facts rather than risks, and in Q2, 3 of 4 paragraphs came from other Items. AMD's four extracted Items total 234{,}195 characters, roughly 58{,}000 tokens at four characters per token, more than Mistral-7B's context window can hold. In Q5 the data step reused the \code{amd\_*} items from Q4, and the report agent received Q4's revenue notes, so a question about tone was answered with revenue growth and never stated the 5/3 sentiment split.

\paragraph{Latency.} Mistral accounted for nearly all recorded time: 9{,}055 of 9{,}057\,ms in Q4, and about 32.5\,s in Q5 (classify 13{,}566\,ms, about 1.7\,s per passage, plus report 18{,}899\,ms). The router step showed 0\,ms although Laya took about 0.3\,s per question in E3, so its time is probably not recorded.

In short, routing was right on 2 of 2 agent questions, validation passed 2 of 2, and 5 of 5 answers gave no advice; but 0 of 2 agent answers were about the company asked about, 2 of 2 contained a mis-identified citation, and 2 of 3 filing answers drifted off the question. None of these is a failure of routing, arithmetic, or live data; all arise at the joins between components.

\begin{table}[!t]
\centering
\footnotesize
\caption{E6: failures the automatic checks missed, and the proposed fix for each.}
\label{tab:e6-errors}
\renewcommand{\arraystretch}{1.15}
\begin{tabularx}{\columnwidth}{@{}L{1.95cm} L{1.9cm} Y@{}}
\toprule
\textbf{Failure} & \textbf{Seen in} & \textbf{Proposed fix} \\
\midrule
Wrong company & Q4, Q5 & check each note's ticker against the question; say ``sample data'' in the answer text; connect XBRL (K2) \\
Citation ID points to the wrong passage & Q4, Q5 (P2) & one passage numbering for prompt and Sources; a claim--passage support check \\
Off-topic content & Q1 (2 of 8 points), Q2 (3 of 4 paragraphs) & send only the Item asked about (1A for risks, 7 for MD\&A) \\
Part of the question unanswered & Q4 (margins), Q5 (5/3 split) & margin notes; pass the sentiment breakdown to the report agent \\
Stale items reused & Q5 (\code{amd\_*}, N1--N3) & key reuse by company, period, filing, and route \\
Unchecked number & Q2 (\$200.97B) & check text figures against XBRL \\
Coarse citations & Q3 (F1--F4 at the end) & one citation per sentence \\
Empty chart & Q5 & a snapshot test for each chart type \\
\bottomrule
\end{tabularx}
\end{table}

\subsection{Shared evidence defeats agreement and shared checks (E8)}
\label{sec:res-e8}

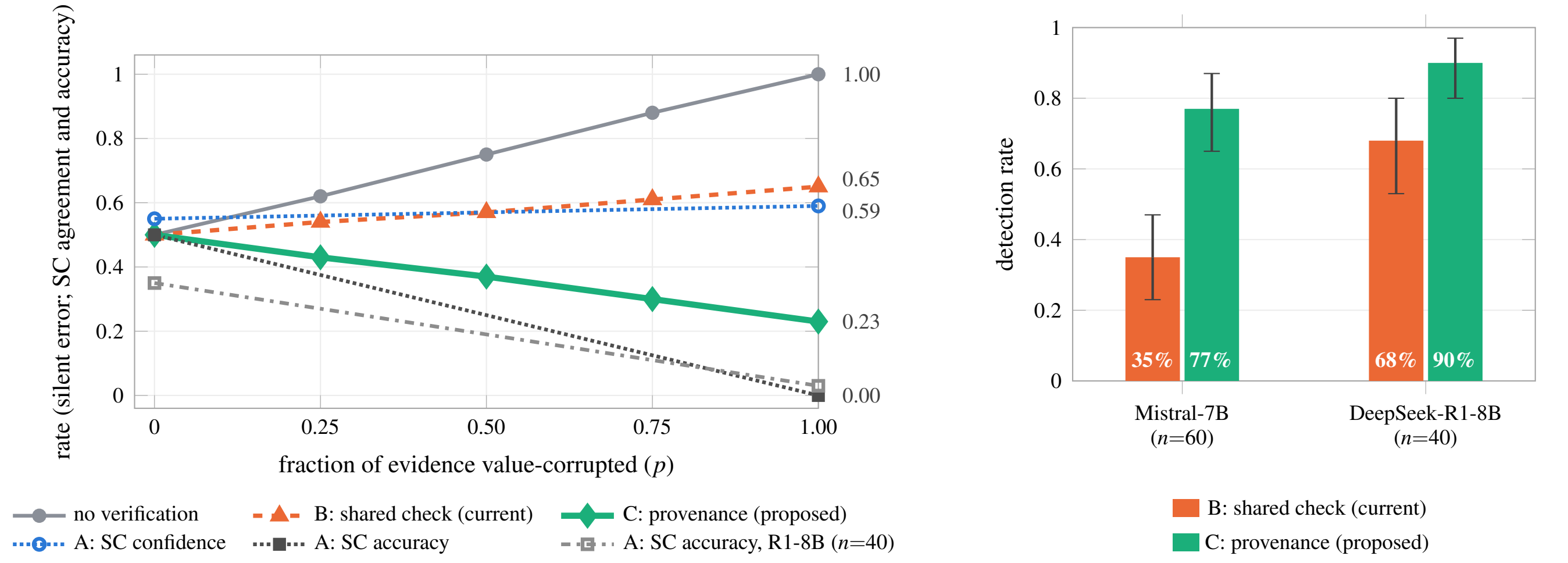
\begin{figure*}[!t]
\centering
\definecolor{nvgray}{HTML}{8A8F98}\definecolor{bcol}{HTML}{EB6834}\definecolor{ccol}{HTML}{1BAF7A}\definecolor{sccol}{HTML}{2A78D6}
\begin{minipage}[b]{0.585\textwidth}
\centering
\begin{tikzpicture}
\begin{axis}[width=0.9\linewidth, height=5.6cm, xmin=-0.03, xmax=1.0, ymin=-0.04, ymax=1.06,
  xtick={0,0.25,0.5,0.75,1}, xticklabels={0,0.25,0.50,0.75,1.00}, ytick={0,0.2,0.4,0.6,0.8,1.0},
  yticklabel style={/pgf/number format/fixed,/pgf/number format/precision=1},
  xlabel={fraction of evidence value-corrupted ($p$)}, ylabel={rate (silent error; SC agreement and accuracy)},
  grid=major, grid style={black!7}, axis line style={black!35}, tick style={black!35},
  tick label style={font=\scriptsize}, label style={font=\footnotesize}, clip=false,
  legend style={font=\scriptsize, draw=none, fill=none, at={(0.47,-0.24)}, anchor=north, legend columns=3, /tikz/every even column/.append style={column sep=6pt}, row sep=-1pt},
  legend cell align=left]
\addplot[color=nvgray, line width=1.2pt, mark=*, mark size=1.7pt] coordinates {(0,0.50)(0.25,0.62)(0.5,0.75)(0.75,0.88)(1,1.00)};
\addlegendentry{no verification}
\addplot[color=bcol, line width=1.4pt, dashed, mark=triangle*, mark size=2.3pt, mark options={solid}] coordinates {(0,0.50)(0.25,0.54)(0.5,0.57)(0.75,0.61)(1,0.65)};
\addlegendentry{B: shared check (current)}
\addplot[color=ccol, line width=2pt, mark=diamond*, mark size=2.6pt] coordinates {(0,0.50)(0.25,0.43)(0.5,0.37)(0.75,0.30)(1,0.23)};
\addlegendentry{C: provenance (proposed)}
\addplot[color=sccol, line width=1.2pt, densely dotted, mark=o, mark size=1.8pt, mark options={solid}] coordinates {(0,0.55)(1,0.59)};
\addlegendentry{A: SC confidence}
\addplot[color=black!70, line width=1.2pt, densely dotted, mark=square*, mark size=1.5pt, mark options={solid}] coordinates {(0,0.50)(1,0.00)};
\addlegendentry{A: SC accuracy}
\addplot[color=black!45, line width=1.2pt, dashdotted, mark=square, mark size=1.6pt, mark options={solid}] coordinates {(0,0.35)(1,0.03)};
\addlegendentry{A: SC accuracy, R1-8B ($n{=}40$)}
\node[font=\scriptsize, anchor=west, text=black!75] at (axis cs:1.02,1.00) {1.00};
\node[font=\scriptsize, anchor=west, text=black!75] at (axis cs:1.02,0.675) {0.65};
\node[font=\scriptsize, anchor=west, text=black!75] at (axis cs:1.02,0.575) {0.59};
\node[font=\scriptsize, anchor=west, text=black!75] at (axis cs:1.02,0.23) {0.23};
\node[font=\scriptsize, anchor=west, text=black!75] at (axis cs:1.02,0.00) {0.00};
\end{axis}
\end{tikzpicture}

{\footnotesize (a) Silent error and self-consistency (SC) vs.\ corrupted fraction}
\end{minipage}\hfill
\begin{minipage}[b]{0.385\textwidth}
\centering
\begin{tikzpicture}
\begin{axis}[width=\linewidth, height=5.6cm, ymin=0, ymax=1.0, ytick={0,0.2,0.4,0.6,0.8,1.0},
  yticklabel style={/pgf/number format/fixed,/pgf/number format/precision=1},
  symbolic x coords={M,R}, xtick={M,R}, xticklabels={Mistral-7B\\($n{=}60$), DeepSeek-R1-8B\\($n{=}40$)},
  xticklabel style={align=center, font=\scriptsize}, enlarge x limits=0.45,
  ylabel={detection rate}, grid=major, grid style={black!7}, xmajorgrids=false,
  axis line style={black!35}, tick style={black!35}, tick label style={font=\scriptsize}, label style={font=\footnotesize},
  ybar=1.5pt, bar width=0.62cm, clip=false,
  legend style={font=\scriptsize, draw=none, fill=none, at={(0.5,-0.30)}, anchor=north, legend columns=1, row sep=1pt},
  legend image code/.code={\draw[#1, draw=none] (0cm,-0.08cm) rectangle (0.3cm,0.12cm);}]
\addplot[fill=bcol, draw=none, error bars/.cd, y dir=both, y explicit, error bar style={black!75, line width=0.8pt}, error mark options={rotate=90, mark size=2.6pt, black!75}] coordinates {(M,0.35) += (0,0.12) -= (0,0.12) (R,0.68) += (0,0.12) -= (0,0.15)};
\addlegendentry{B: shared check (current)}
\addplot[fill=ccol, draw=none, error bars/.cd, y dir=both, y explicit, error bar style={black!75, line width=0.8pt}, error mark options={rotate=90, mark size=2.6pt, black!75}] coordinates {(M,0.77) += (0,0.10) -= (0,0.12) (R,0.90) += (0,0.07) -= (0,0.10)};
\addlegendentry{C: provenance (proposed)}
\node[font=\scriptsize\bfseries, text=white] at ([xshift=-0.33cm]axis cs:M,0.06) {35\%};
\node[font=\scriptsize\bfseries, text=white] at ([xshift=0.33cm]axis cs:M,0.06) {77\%};
\node[font=\scriptsize\bfseries, text=white] at ([xshift=-0.33cm]axis cs:R,0.06) {68\%};
\node[font=\scriptsize\bfseries, text=white] at ([xshift=0.33cm]axis cs:R,0.06) {90\%};
\end{axis}
\end{tikzpicture}

{\footnotesize (b) Detection rate on corrupted evidence, 95\% bootstrap CI}
\end{minipage}
\caption{E8 on FinRank (Mistral-7B, $n = 60$; DeepSeek-R1-8B, $n = 40$). (a)~As more evidence is value-corrupted, silent error (a wrong answer that is not flagged) rises to 1.00 without verification and to 0.65 with the shared-evidence check, but falls to 0.23 with independent provenance. Self-consistency, measured on clean ($p = 0$) and fully corrupted ($p = 1$) evidence, stays confident (agreement 0.55 to 0.59) while its accuracy falls from 0.50 to 0.00; DeepSeek-R1-8B's accuracy (gray dash-dot, $n = 40$) also collapses, from 0.35 to 0.03. (b)~On corrupted evidence, the shared-evidence check flags 35\% of Mistral-7B's answers [0.23, 0.47] and independent provenance 77\% [0.65, 0.87] (paired exact McNemar $p = 5.96\times10^{-8}$; C caught 25 answers that B missed, B none that C missed); for DeepSeek-R1-8B, 68\% [0.53, 0.80] against 90\% [0.80, 0.97] ($p = 0.003906$; 9 against 0). Check C consults the uncorrupted gold passage, so its detection rates are an upper bound.}
\label{fig:e8}
\end{figure*}

\begin{table}[!t]
\centering
\footnotesize
\setlength{\tabcolsep}{5pt}
\caption{E8: silent-error rate as the fraction $p$ of value-corrupted evidence grows (Mistral-7B, $n = 60$). Rates at intermediate $p$ equal the mixture of the clean ($p = 0$) and fully corrupted ($p = 1$) rates.}
\label{tab:e8}
\begin{tabular}{@{}cccc@{}}
\toprule
$p$ & \textbf{No verification} & \textbf{B: shared check} & \textbf{C: provenance} \\
\midrule
0.00 & 0.50 & 0.50 & 0.50 \\
0.25 & 0.62 & 0.54 & 0.43 \\
0.50 & 0.75 & 0.57 & 0.37 \\
0.75 & 0.88 & 0.61 & 0.30 \\
1.00 & 1.00 & 0.65 & \textbf{0.23} \\
\bottomrule
\end{tabular}
\end{table}

\paragraph{Self-consistency gave false confidence.} With clean evidence, the five samples agreed 0.55 of the time and the majority figure was right 0.50 of the time. With corrupted evidence, agreement stayed flat at 0.59 while accuracy fell to 0.00 (\cref{fig:e8}a): the samples agreed with one another because they read the same wrong number.

\paragraph{A shared-evidence check was not enough.} On corrupted evidence, the shared-evidence check flagged only 0.35 of the answers (95\% CI [0.23, 0.47]), because the wrong figure is present in the very passage it checks. Independent provenance flagged 0.77 (95\% CI [0.65, 0.87]), and the intervals do not overlap (\cref{fig:e8}b). The paired comparison is one-sided: C caught 25 answers that B missed, and B caught none that C missed (exact McNemar $p = 5.96\times10^{-8}$), so C's detections are a superset of B's.

\paragraph{Silent error grows with shared corruption.} As the corrupted fraction rises from 0 to 1, silent error rises from 0.50 to 1.00 without verification and from 0.50 to 0.65 with the shared check, but falls from 0.50 to 0.23 with independent provenance (\cref{tab:e8}). At full corruption, independent provenance cuts silent error from 0.65 to 0.23. The 0.50 at $p = 0$ is the clean-context noise floor, the same for all conditions.

\begin{table}[!t]
\centering
\footnotesize
\setlength{\tabcolsep}{3pt}
\caption{E8: a standard and a reasoning LLM under the same value corruption on FinRank. SC = self-consistency (agr.\ = agreement of five samples; accuracy of the majority figure). Detection is on corrupted evidence, with 95\% bootstrap intervals.}
\label{tab:e8-models}
\begin{tabular}{@{}lcc@{}}
\toprule
 & \textbf{Mistral-7B} & \textbf{DeepSeek-R1-8B} \\
 & $n = 60$ & $n = 40$ \\
\midrule
Clean: SC agr.\ / accuracy & 0.55 / 0.50 & 0.26 / 0.35 \\
Corrupted: SC agr.\ / accuracy & 0.59 / 0.00 & 0.20 / 0.03 \\
\midrule
B: shared-check detection & 0.35 [0.23, 0.47] & 0.68 [0.53, 0.80] \\
C: provenance detection & \textbf{0.77} [0.65, 0.87] & \textbf{0.90} [0.80, 0.97] \\
McNemar C vs.\ B ($b$ / $c$) & 25 / 0 & 9 / 0 \\
\quad exact $p$ & $5.96\times10^{-8}$ & 0.003906 \\
\bottomrule
\end{tabular}
\end{table}

\paragraph{Reasoning does not resist corruption.} We repeated E8 with DeepSeek-R1-8B on 40 items (\cref{tab:e8-models}). Reasoning did not recover the true value: with clean evidence, R1's majority figure was right 0.35 of the time, and with corrupted evidence 0.03. Its samples agreed less often than Mistral's in both conditions (0.26 and 0.20), reflecting noisier figure extraction, so the two models are not identical: R1's agreement fell slightly under corruption, but far less than its accuracy did. Independent provenance again dominated the shared check. The shared check flagged 0.68 of R1's corrupted-evidence answers [0.53, 0.80], more than for Mistral but still missing about a third, while independent provenance flagged 0.90 [0.80, 0.97]; C caught 9 answers that B missed, and B caught none that C missed (exact McNemar $p = 0.003906$). The core result therefore holds for both a standard and a reasoning LLM.

\paragraph{The dangerous error is a plausible wrong value.} Replacing the gold passage with FinRank's hard negatives did not reproduce the failure: given these confusable passages, the model mostly abstained rather than stating a confident wrong figure. The dangerous case is a wrong value inside a plausible, on-topic passage. This is also the mechanism behind E6: \sys{}'s validator compares a draft with the notes it was written from, so AMD sample data presented under NVIDIA's name passed, just as a corrupted figure passes check B.

\section{Discussion}
\label{sec:discussion}

\paragraph{Different models for different jobs.} The same generative model that labeled sentiment best (0.93) was the worst router (0.52) and a poor calculator (29\%). Assigning each job to the component that does it best produced the highest-scoring system (0.88, in-sample) and removed invented numbers and invented prices. Newer LLMs narrowed the arithmetic gap (72\%, 98\%) but stayed inexact and orders of magnitude slower than code, so their strengths are better used for writing. Nor did scale fix routing: a 12B Mistral decided no better than the 7B and was slower. Explicit reasoning helped more: DeepSeek-R1-8B's overall gain was significant on a matched run (79 against 68 of 100; $p = 0.0192$), and in its fully scored run it raised routing-intent accuracy from 0.5167 to 0.6667, which is consistent with routing errors coming partly from a lack of step-by-step deliberation rather than from capacity, although R1 also differs from Mistral in base model, training, and decoding. Yet R1 took about 9.9 times longer per decision and still trailed Laya, which, with about 421M parameters, routed 0.85 of the same questions. A dedicated typed-decision model therefore remains the right router. A reasoning model fits better in latency-tolerant stages, such as planning a multi-step analysis or drafting a longer research report, where its output is still checked by code and every number still comes from Python. E7 measures the LLMs only as decision-makers, however; a larger or reasoning model may still write better answers, which we did not test.

\paragraph{Confidence is per decision.} A router that is confidently wrong (0.97 on a wrong route) cannot be gated by its own confidence, and a cascade or agreement rule helps only when the fallback is stronger than the primary. Laya's 0.87 is a confidence in the route; the case study shows that it says nothing about the data that follows. E8 adds that agreement among samples is not evidence of correctness: when every sample reads the same wrong number, they agree with it. Confidence should be reported and used per decision (route, advice, label, citation) and never read as a score for the whole answer.

\paragraph{Verify against an independent source.} The validator guaranteed that every displayed number came from code, and it held. But any check that reads the same evidence as the writer inherits that evidence's errors. In E8, a wrong figure inside a plausible passage left self-consistency confident (0.59 agreement at 0.00 accuracy) and slipped past the shared-evidence check in 65\% of cases, while a check against the original source caught 77\% and missed nothing the shared check caught. Reasoning does not remove the problem: DeepSeek-R1-8B did not recover the true value under corruption (accuracy 0.03), while provenance still caught 90\% of its errors against 68\% for the shared check. Reasoning helped when the evidence was right (E7) but could not correct evidence that was wrong. E6 showed the same mechanism in the live system. Trustworthy answers therefore need verification that is independent of the evidence path, with each figure re-derived from the filing or its XBRL value and fetched separately from the copy the writer saw. They also need three cheap checks from E6: the \emph{entity} of every note matches the question, every cited passage \emph{supports} its sentence, and every part of the question is \emph{covered}.

\paragraph{Narrow before writing, and key what you reuse.} The writer used what it was given, not what was asked: four Items produced a summary of four Items, and leftover revenue notes produced a revenue answer to a tone question. Section-first retrieval and route-scoped prompts are the remedy, and recycled memory must be keyed to what it describes.

\paragraph{Engineering is part of accuracy.} The most consequential errors were not model errors: a table-of-contents stub that hid the risk factors, memory that leaked across companies, substring matching that refused harmless questions, a stale ``latest'' filing, and a passage ID that pointed to the wrong text. Explainable traces made each of them visible.

\section{Limitations and Ethics}
\label{sec:limits}
\label{sec:ethics}
Our test sets are small and partly synthetic (100 decision items, 100 clean calculations, 10 tickers, 50 retrieval questions, a six-question embedder pilot, and five case-study questions); D1's gold labels await full human sign-off; E1 and E7 report all 100 items without a held-out split, so the hybrid's 0.88 is in-sample; retrieval correctness is judged by patterns; each configuration ran once, except DeepSeek-R1-8B in E7, which ran twice; and latencies come from different hardware. gpt-5 was scored on 50 problems. We have not yet measured answer-level faithfulness or citation accuracy, the agent pipeline's numbers and sentiment routes still use sample data (K2), and all companies are large U.S.\ technology firms. The case study identifies failure types, not their rates. In E8, clean-context accuracy is 0.50, a noise floor from extracting figures out of raw tabular passages; it affects all conditions equally and sets the silent-error baseline at $p = 0$. E8 uses $n = 60$ items from the verbatim-extractable subset, a single corruption-magnitude band, and one 4-bit model; its accuracy and detection use numeric-grounding heuristics (significant-figure matching), not full answer semantics; and it is a single run, with self-consistency sampled at temperature 0.7. The DeepSeek-R1-8B repetition uses 40 items rather than 60, so its rates are not directly comparable with Mistral-7B's item for item. E8's independent-provenance check compares against the uncorrupted gold passage, so its detection rates are an upper bound, and it assumes that a correct independent source is available; in \sys{} that source would be the filing or XBRL value itself. E7 shares D1's limits (a small evaluation set drawn from three semiconductor companies, with a draft gold standard pending correction), and the router has no explicit out-of-scope branch, although D1 contains five out-of-scope questions. E7 measures each model only as a classifier and decision-maker, not the quality of its generated answers, which a larger model could still affect. Its models differ in base model and training as well as in size or reasoning. R1 and the Mistral models use different decoding budgets by necessity (reasoning versus a single word); this is required for a fair test of a reasoning model, but it means the latency comparison reflects each model's intended operating mode, not identical token budgets. R1 was sampled at temperature 0.6, and its two runs scored 0.77 and 0.79; only the second saved per-item results, so the paired test and the sub-task metrics come from different runs, and a single run that saves per-item results would remove this split. All models are 4-bit quantized through Ollama, and results may differ at full precision. We did not run a paired significance test between Mistral-7B and NeMo-12B, so that conclusion of no improvement rests on identical point accuracy and fully overlapping bootstrap intervals. \sys{} is a research tool, not an investment adviser: advice is refused before any model runs, answers carry citations, sample data must be labeled in the answer text and not only on the chart, filing data is accessed under the SEC's fair-access guidelines, and news is labeled untrusted to limit prompt injection through MCP. A user study would require institutional review. An AI writing assistant helped draft and edit the text and check the references; the authors designed and ran all experiments, checked every reported number against their results, and take full responsibility for the content.

\section{Conclusion and Future Work}
\label{sec:conclusion}
\label{sec:future}
We built \sys{} around one question (which component should do each job?) and measured the answer live. Six specialized agents divide the work: Laya routes (0.85 vs.\ 0.52 for Mistral; a 12B Mistral decided no better, and a reasoning model reached 0.6667 on routing intent at about 9.9 times the latency), Mistral labels sentiment (0.93) and writes, Python computes (100\% vs.\ 29--98\% for three LLMs), live data fixes prices (210\% error to 0.00\%), and hybrid retrieval finds the evidence (Hit@5 76\%); a per-task hybrid scored highest (0.88, an in-sample estimate). The live case study shows the next step: correct routing and verified numbers did not prevent answers about the wrong company or citations to the wrong passage. On FinRank, the reason is clear: checks that share the writer's evidence pass its errors (35\% caught), while independent provenance catches most of them (77\%), and the ordering is the same for a reasoning model (DeepSeek-R1-8B: 90\% against 68\%). We will (i) add independent-provenance checks against the original filing and XBRL values, together with entity, citation-support, and coverage checks scored with ALCE-style citation metrics \citep{gao2023alce,es2024ragas}, re-ask the same five questions as a regression test, and extend E8 to more corruption magnitudes, more items, and full-precision models; (ii) finish the XBRL and retrieval hookups, key memory reuse, and send the writer only the Item asked about; (iii) evaluate on FinanceBench \citep{islam2023financebench}, FinQA, TAT-QA, and DocFinQA against LLM-only, RAG, and program-of-thought baselines, adopt Qwen3-Embedding-0.6B if it holds up in the full harness, and compare FinBERT on sentiment; evaluate DeepSeek-R1 in latency-tolerant planning and report stages, with every number still computed in code; (iv) add a ``what changed'' view between consecutive filings, motivated by evidence that such changes are informative \citep{cohen2020lazy}; and (v) package the system to run on a single commodity GPU (Ministral~3~8B already ran on a T4) and conduct a user study with finance students under institutional review. Laya decides, Python computes, tools fetch, Mistral writes, and code checks: measured, not assumed.

\section*{Acknowledgments}
We thank the CMPE 259 instructor and teaching staff at San Jos\'e State University for feedback that shaped the separation of stored and real-time data and the intent-based routing. This work used the Bridges-2 system at the Pittsburgh Supercomputing Center \citep{brown2021bridges2} and Jetstream2 at Indiana University \citep{hancock2021jetstream2} through allocation CIS261572 from the Advanced Cyberinfrastructure Coordination Ecosystem: Services \& Support (ACCESS) program \citep{boerner2023access},
which is supported by U.S.\ National Science Foundation grants \#2138259, \#2138286, \#2138307, \#2137603, and \#2138296.

\section*{Code and Data Availability}
\begingroup\raggedright
Code and evaluation data are available at \url{https://github.com/SJSU-CMPE259/AI-Stock-Research-and-Earnings-Assistant}: the evaluation harness and labeled sets in \nolinkurl{ACSESS/finsight/agent_core/eval/}, the experiment scripts and findings in \nolinkurl{ACSESS/finsight/agent_core/experiments/}, the retrieval evaluation in \nolinkurl{src/finsight/rag/eval.py}, and the MCP server tests in \nolinkurl{ACSESS/finsight/mcp_server/tests/} (branch \nolinkurl{feat/mcp-server}).\par
\endgroup

\begingroup
\footnotesize
\setlength{\bibsep}{1.5pt plus 0.5pt}
\bibliographystyle{unsrtnat}
\bibliography{references}
\endgroup

\end{document}